\pdfoutput=1
\documentclass[11pt]{article}

\usepackage[]{emnlp2021}

\usepackage{times}
\usepackage{latexsym}

\usepackage[T1]{fontenc}
\usepackage[utf8]{inputenc}

\usepackage{microtype}

\usepackage{graphicx, float}
\usepackage{natbib}
\renewcommand{\cite}[1]{\citep{#1}}

\usepackage[ruled,vlined]{algorithm2e}
\usepackage{afterpage}
\usepackage{float}
\usepackage{subfig}

\usepackage{booktabs}
\usepackage{longtable}
\usepackage{tikz}
\usepackage{amsmath}
\usepackage{multirow}
\usepackage{todonotes}

\usepackage{booktabs}
\usepackage{array}
\usepackage{colortbl}

\usepackage{subfig}

\usepackage{todonotes}
\usepackage{hyperref}
\title{Translation Transformers Rediscover Inherent Data Domains}

\author{Maksym Del\thanks{\phantom{*}Equal contribution} , Elizaveta Korotkova\footnotemark[1] , Mark Fishel \\
  Institute of Computer Science \\
  University of Tartu, Estonia \\
  \texttt{\{maksim,lisa\_k,mark\}@tartunlp.ai} }

\begin{document}
\maketitle
\begin{abstract}

Many works proposed methods to improve the performance of Neural Machine Translation (NMT) models in a domain/multi-domain adaptation scenario. However, an understanding of how NMT baselines represent text domain information internally is still lacking. Here we analyze the sentence representations learned by NMT Transformers and show that these explicitly include the information on text domains, even after only seeing the input sentences without domains labels. Furthermore, we show that this internal information is enough to cluster sentences by their underlying domains without supervision. We show that NMT models produce clusters better aligned to the actual domains compared to pre-trained language models (LMs). Notably, when computed on document-level, NMT cluster-to-domain correspondence nears 100\%. We use these findings together with an approach to NMT domain adaptation using automatically extracted domains. Whereas previous work relied on external LMs for text clustering, we propose re-using the NMT model as a source of unsupervised clusters. We perform an extensive experimental study comparing two approaches across two data scenarios, three language pairs, and both sentence-level and document-level clustering, showing equal or significantly superior performance compared to LMs.

\end{abstract}

\section{Introduction}
%TODO: remove references to the Domain Control scenario  

\begin{figure*}[t!]
\includegraphics[width=\textwidth]{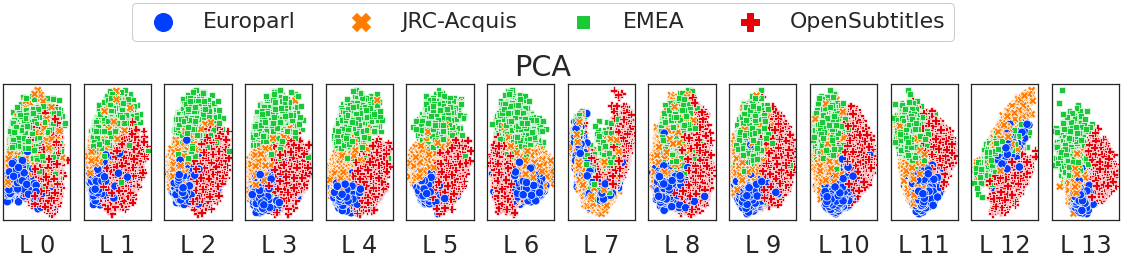}

\centering
\caption{%t-SNE and 
PCA plots of sentence representations extracted from all layers of the 60th checkpoint of the trained baseline NMT model. Representations are computed with English sentences. The dots, denoting sentences, are colored according to the domain the corresponding sentences come from. The model learns to distinguish between domains in its hidden space, despite not being explicitly provided with any information about domains. L0 corresponds to fixed encoder embeddings, L1--L6 are encoder layers' representations, L7 shows fixed decoder embeddings and L8--L13 -- the decoder layers' representations. The figure shows that representations from the same domain cluster together.}

\label{fig:domains-tsne-pca}
\end{figure*}

Neural machine translation \citep[NMT,][]{Bahdanau2015NeuralMT,vaswani2017attention} heavily depends on training data and the text domains covered in it. Full-scale NMT Transfomer models  \cite{vaswani2017attention}  are usually trained on a mix of corpora from several domains \citep{barrault-etal-2020-findings}. However, the field lacks an understanding of how these NMT models represent the training data domains in their inner vector spaces.

This paper consists of two main parts. First, we analyze domain representations learned by the NMT Transformer. We consider sentence-level as well as document-\-level representations via mean pooling of token contextual embeddings. Our analysis shows that NMT models %keep domains separately
explicitly learn to include the domain information
in their representational spaces across layers. 
Furthermore, we show that text representations preserve enough domain-specific information to reveal the underlying domains with Principal Component Analysis and k-means clustering without supervision. In the case of document-level clustering, the result of k-means matches the original corpora almost perfectly. In the case of sentence-level clustering, we observe some deviation between automatic clusters and the original corpora that the sentences belong to, showing corpus heterogeneity on the sentence level.

\citet{aharoni-goldberg-2020-unsupervised} previously revealed that a similar property exists in pre-trained language models (LMs). We compare LMs with NMT Transformers in how well we can extract unsupervised domain clusters from them and show the superiority of NMT models.  

In the second part of the paper, we show how to effectively utilize our analysis to improve an existing approach to NMT domain adaptation which uses automatically extracted domains \cite{tars18, currey-etal-2020-distilling}. This method targets the case when training domain labels are not precise \citep[e.g.][]{currey-etal-2020-distilling} or missing overall, as in case of heterogeneous corpora \citep[e.g. Paracrawl,][]{paracrawl}. This framework has so far been used with external models for clustering, which automatically makes us rely on clusters not necessarily aligned with the objectives of translation or target data domains.

We propose exploiting clusters extracted from the NMT baseline (already trained as a part of the framework) to improve translation quality without relying on external language models. We test our proposal empirically, covering three language pairs and two data settings: a mix of corpora with known domain labels and a heterogeneous corpus without such labels. We show that fine-tuning the NMT models to the automatically discovered clusters on average matches or surpasses tuning to the original corpus labels (when available) and deep LM-based clusters.
% Fine-tuning to document-level clusters shows same or than sentence-level clustering, based on comparisons of both BERT-based and NMT-based representations.

%The novelty of our research is summarized in the following contributions:
Our contributions are thus two-fold:\footnote{We release our code at \url{https://github.com/TartuNLP/inherent-domains-wmt21}}
\begin{itemize}
    %\item we perform extensive analysis of the NMT encoder's ability to automatically discover inherent text domains (Section~\ref{sec:analysis});
    \item we analyze the NMT encoder's representations, showing their ability to automatically discover inherent text domains and cluster unlabelled corpora, testing both sentence-level and document-level  representations (Section~\ref{sec:analysis});
    \item we utilize findings from our analysis to improve an existing Automatic Domains for NMT approach (Section \ref{sec:framework}) and perform an extensive experimental study, 
    showing the superiority of our method (Section \ref{sec:experiments});
    % \item we introduce an approach to text clustering for NMT based on this analysis, on both sentence and document-level, showing that the discovered representation clusters can lead to translation quality improvements (Sections~\ref{sec:framework} and \ref{sec:experiments});
\end{itemize}

\section{Related Work}

\citet{aharoni-goldberg-2020-unsupervised} found that BERT \citep{devlin2019bert} produces meaningful unsupervised domain clusters and used this finding for NMT data selection. In this work we analyse (sentence-level and document-level) hidden representations produced by a baseline NMT model and find that it learns superior unsupervised clusters by itself.

In NMT, domain-specific information on the word level was recently analyzed by \citet{jiang-etal-2020-multi} in the context of domain mixing in a joint modular multi-domain NMT system. They found that representations contain domain-specific information related to the multiple domains in different proportions on the word level. We analyze representation on the sentence and document level, revealing that domain-specific information in representations converges to the one specific domain with a broader context.  

\citet{currey-etal-2020-distilling}
%, inspired by \citet{aharoni-goldberg-2020-unsupervised}, 
used contextual embeddings and mean-pooled representation clustering for domain adaptation. We compare our approach to \citet{currey-etal-2020-distilling}, however in their case the representations were extracted from multilingual BERT (mBERT). We cluster based on the NMT encoder's representations directly and also experiment with document-level representations in addition to sentence-level ones.

Before \citet{currey-etal-2020-distilling}, the automatic domains framework has been used in NMT only with external models for clustering as well. \citet{tars18} used fixed embeddings from FastText \citep{bojanowski-etal-2017-enriching} for clustering mean-pooled sentence representations and then either tuning NMT systems to these clusters or supplying the cluster identity to the NMT system as additional input for multi-domain translation.

\section{Analysis} 
\label{sec:analysis}

In this section, we perform an analysis of inherent domain representations in translation transformers. We reveal how well the domain-specific information in text representations is preserved in NMT models. We focus on "out-of-the-box" NMT systems without any changes and explore the extent to which we can use their internal representations to match the original text domains using Principal Component Analysis (PCA) and k-means clustering. We also measure the effect of using broader document-level representations. 

Additionally, we compare NMT representations to the ones extracted from a pre-trained language model, for which \citet{aharoni-goldberg-2020-unsupervised} revealed a high degree of domain-specific information.

\subsection{Models and Data}
\label{sec:analysis:setup}
In our analysis, we start by following \citet{currey-etal-2020-distilling} and similarly to them use a multilingual LM \citep[XLM-R, ][]{conneau-etal-2020-unsupervised} to obtain clusters. XLM-R is a multilingual masked language modeling transformer covering 100 languages.

We then train \textit{Transformer-base} \citep{NIPS2017_3f5ee243} NMT models, which have $\sim$97M parameters each. We train the models on parallel data covering four corpora/text domains: parliament speeches \citep[Europarl,][]{koehn-europarl}, medical \citep[EMEA,][]{OPUS},  subtitles \citep[OpenSubtitles,][]{opensubs} and legal \citep[JRC-Acquis,][]{steinberger-jrc-acquis}. We sub-sampled the larger corpora in order to balance the size of training data across domains. The NMT models were trained for 60 epochs. A detailed description of the setup, models, and data is provided in Appendix \ref{app:experiments_setup}.

We focus on sentence-level and document-level representations, and two language pairs: English$\rightarrow$Estonian (EN-ET) and German$\rightarrow$English (DE-EN).

\subsection{Dimensionality Reduction} 
We start by unsupervised dimensionality reduction using PCA to visualize domain placement. We take the development set data, extract token embeddings from each model's layer, and average them to obtain sentence representations. Then we apply cosine-based PCA and t-SNE dimensionality reduction to the representations to visualize the data in a 2D space, and post factum color each data point (sentence) according to its corresponding domain. We show the resulting visualizations in Figure \ref{fig:domains-tsne-pca} (best viewed in color) for ET-EN (and in Figure \ref{fig:domains-tsne-only} for t-SNE in the Appendix \ref{app:analysis}, which mirrors the PCA result).

Figure \ref{fig:domains-tsne-pca}  shows that NMT partitions the domains quite well at all encoder hidden layers and deep decoder layers. Encoder layer 0 corresponds to the fixed embeddings, and the latent space is not well partitioned there yet; however, as we go deeper into the network, the separation increases. Layer 7 is the decoder's embedding layer, and there the same logic applies. While the encoder learns to partition the hidden space based on domains from scratch, the decoder has access to the encoder hidden states via encoder-decoder attention, which might simplify its task. 

In summary, Figure \ref{fig:domains-tsne-pca} is our initial evidence that the NMT encoder places the domains separably.

\subsection{Clustering}
\label{sec:analysis:method}

Our primary method, however, is unsupervised k-means clustering. We consider four data clustering setups: sentence-level XLM-R clusters, sentence-level NMT clusters, document-level XLM-R clusters, and document-level NMT clusters. The first one is the baseline clustering approach investigated by \citet{aharoni-goldberg-2020-unsupervised} while the remaining three are our original contributions.

\subsubsection{Per-layer Clustering Purity}
\label{sec:analysis:results-and-discussion}

\paragraph{Metric} In our analysis, we estimate how well the NMT model preserves domain-specific information in its internal text representations. To do that, we measure the goodness-of-fit between unsupervised clusters and oracle domains. Specifically, we follow \citet{aharoni-goldberg-2020-unsupervised} and use the \textit{clustering purity} metric. To compute clustering purity, we align domains and clusters by the highest overlap in numbers of sentences. The number of overlapping data points for each cluster-domain pair gives us the number of 'correctly predicted' examples. Then, the sum of all 'correctly predicted' examples divided by the total number of examples will be the clustering purity score.

\paragraph{Embedding and Clustering} We first take the concatenation of a small subset of sentences (3k) from each of the four domains and try to partition them into four clusters based on the representations from each layer of XLM-R and NMT Transformer. We only use source sentences since we do not have targets at runtime in NMT. Specifically, we follow the steps below for each layer of each of the two models:
\begin{enumerate}
    \item For each sentence in the dataset, we extract contextualized token embeddings from a layer of the model.  
    \item We use the average of contextualized token embeddings as sentence representations. 
    \item We apply k-means clustering to sentence representations to assign a cluster label to each sentence. 
    \item We compute clustering purity for predicted labels and oracle domains.
\end{enumerate}
We perform ten random restarts of k-means clustering, selecting the iteration with the smallest within-cluster variance. 

\begin{figure}%
    \centering
\includegraphics[width=0.5\textwidth]{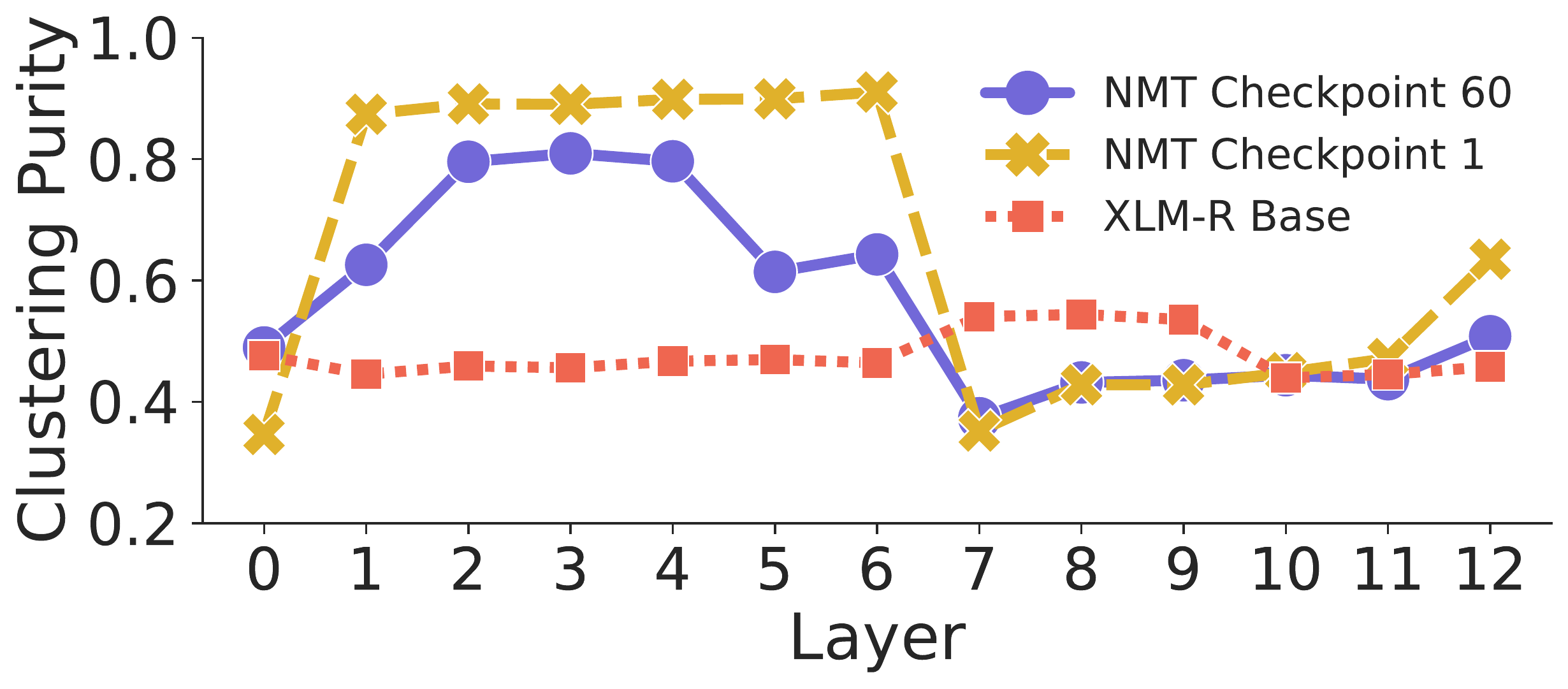} %
\caption{Sentence-level clustering purity between clusters obtained with k-means over 3k of EN-ET development set sentences and actual data domains. Representations extracted from XLM-R, NMT Baseline after epoch 1, and NMT Baseline after training has finished (epoch 60). While XLM-R is relatively poor in its ability to rediscover original domains, representations extracted from the trained NMT model largely outperform it at layers 1-6. Layers 1-6 are the encoder and 7-12 the decoder. The results are for the best clustering (with least variance) over 10 k-means runs. }%
    \label{fig:purity-domains-clusters}%
\end{figure}

\paragraph{Results} Figure \ref{fig:purity-domains-clusters} shows per-layer clustering purity computed for sentence representations for XLM-R and two NMT Baseline checkpoints (after the first epoch and after the 60th epoch of training). Figure \ref{fig:purity-domains-clusters} shows that NMT surpasses the language model in its ability to rediscover domains. About 3.5x higher performance at the encoder layers shows that the encoder is the part that learned to be very aware of the input domains (in an unsupervised way). Figure \ref{fig:purity-domains-clusters} also shows that the checkpoint saved after the 1st training epoch rediscovers clusters slightly better then 60th checkpoint. However, this does not suggest that an NMT model should be trained for one epoch since the translation quality is suboptimal early on. Instead, we assume that the model quickly learns domain-specific information (perhaps due to the common lexical statistics) and then slightly "moves away" towards a higher level of abstraction as training progresses. This abstraction is necessary to successfully learn a task as complex as NMT.

\begin{table}[]
\setlength{\tabcolsep}{3.5pt} % Default value: 6pt
\renewcommand{\arraystretch}{1} % Default value: 1
\small
\centering
\begin{tabular}{@{}lllllll@{}}
\toprule
 & \multicolumn{3}{c}{EN-ET} & \multicolumn{3}{c}{DE-EN} \\ 
\cmidrule(lr){2-4} \cmidrule(lr){5-7}
 & train & dev & test & train & dev & test \\  \midrule
sentence \\
\hspace{3mm} XLM-R & 53.47 & 52.9 & 50.07 & 44.04 & 49.2 & 48.6 \\
\hspace{3mm} NMT  & 67.21 & 72.56 & 70.7 & 66.32 & 70.02 & 72.28\\
document \\
\hspace{3mm} XLM-R & 85.77 & 72.89 & 70.14 & 97.64 & 91.74 & 95.23\\
\hspace{3mm} NMT   & 99.61 & 100.0 & 99.1 & 99.21 & 97.58 & 99.78\\
\bottomrule
\end{tabular}
\caption{Clustering purity. We trained the NMT model on about 2m EN-ET or DE-EN sentences from multiple corpora and used pre-trained XLM-R Base model. Based on the Figure \ref{fig:purity-domains-clusters}, we used the 4th layer to extract source representations from the NMT model and 8th layer for XLM-R. The results are for the best clustering (with least variance) over 10 k-means runs. Both NMT and XLM-R rediscover inherent data domains when document level representations are used, and seem to produce more customized separations when clustered based on sentence-level. NMT tends to be better at rediscovery. }  % Our approach with \textit{document level} clustering, where \textit{NMT model} is used to extract representations, outperforms baseline and all other clustering approaches when the number of clusters is high enough.}
\label{tab:purity-table}
\end{table}

\subsubsection{Large-scale Clustering}
\label{sec:analysis:analysis}

\begin{figure*}%
    \centering
    \subfloat[\centering EN-ET Train]{ {\includegraphics[width=\textwidth]{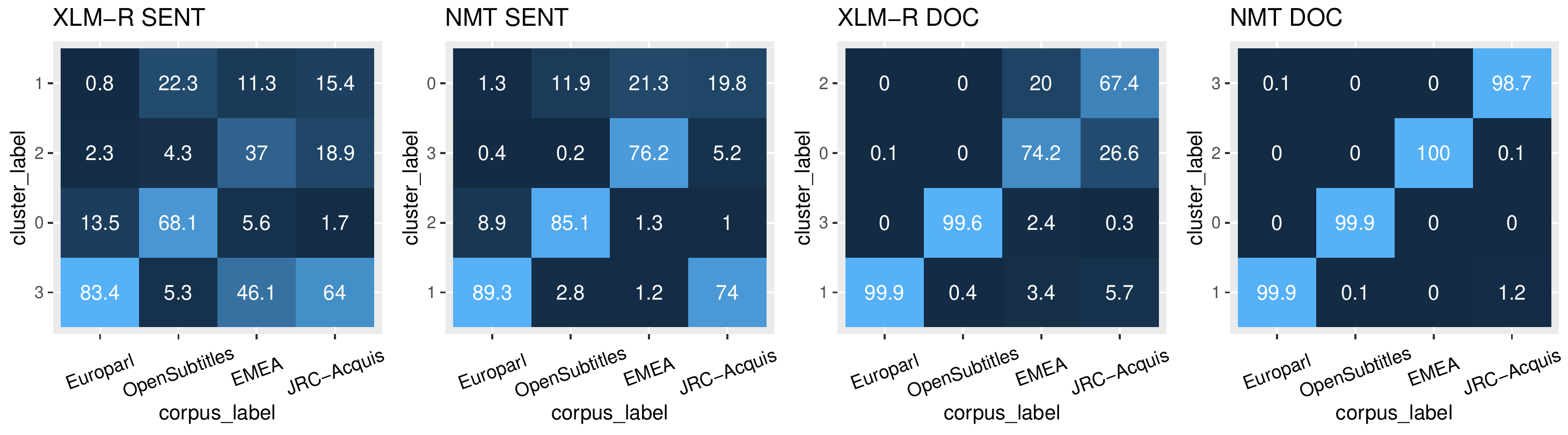} }}%
    \qquad
    \subfloat[\centering EN-ET Test]{ {\includegraphics[width=\textwidth]{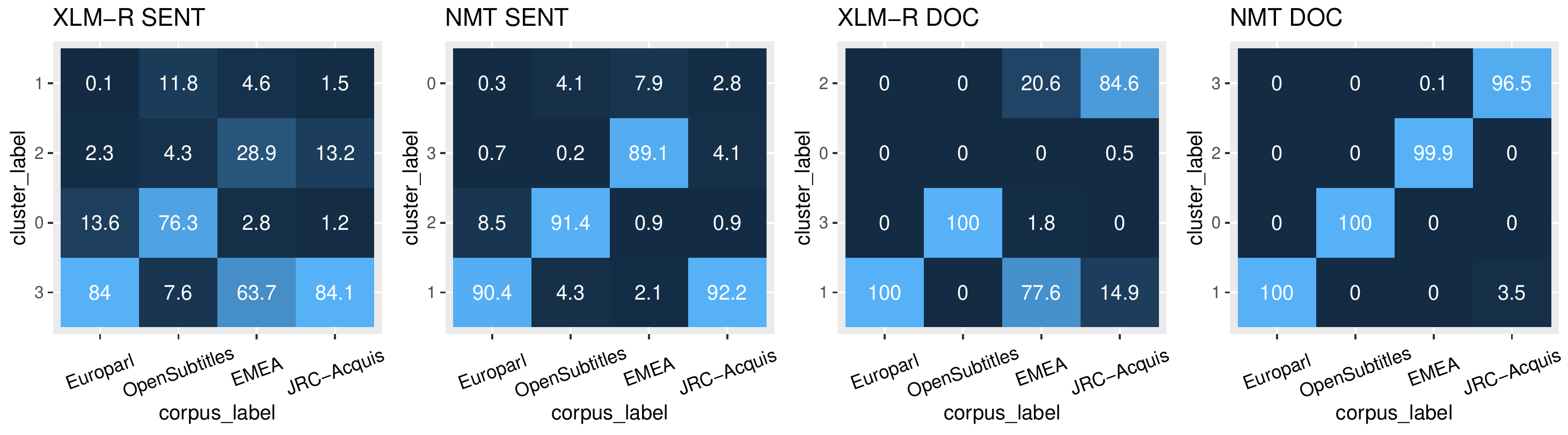} }}%
    % \qquad
    % \subfloat[\centering DE-EN Train]{ {\includegraphics[width=\textwidth]{figures/conf_table-de-en-train.pdf} }}%
    % \qquad
    % \subfloat[\centering DE-EN Test]{ {\includegraphics[width=\textwidth]{figures/conf_table-de-en-test.pdf} }}%
    \caption{Corpus-cluster confusion tables for about 2M sentences for EN-ET for (a) the training and (b) test sets. The numbers are percentages for each domain (column). NMT document clusters almost perfectly match original data domains. On the sentence level, however, both NMT and XLMR learn a more customized notion of clusters, with NMT being more aligned with original domains.}%
    \label{fig:conf_tables}%
\end{figure*}

Next, we repeat the same steps for the entire training dataset and include a second language pair. Specifically, we pick one of the best performing layers (4th for NMT and 7th for XLM-R) based on the experiment above and use it to cluster the training part of the multi-domain machine translation training set (about 500k examples per domain, 2M in total). We then predict cluster labels for the training examples and use the same model to cluster unseen examples from the development and test set.

We also extend our analysis to the document-level scenario. Specifically, we average over sentence representations to get document embeddings and cluster-based on them. Then, we assign the predicted label for each document to each sentence in that document.\footnote{Sentence pairs coming from the same XML file were considered to belong to the same document. The training, development and test sets in all experiments were constructed in such a way that a document is always included in one set in its entirety.}

\paragraph{Results} We present large-scale clustering confusion matrices in Figure \ref{fig:conf_tables} and clustering purity in Table \ref{tab:purity-table}. These show that sentence-level NMT is generally better than sentence-level XLM-R at rediscovering domains. However, they both show quite modest results for both language pairs. At the same time, document-level clusters are much better at rediscovering original domains. 

The reason for that might be that sentence-level clustering largely relies on the more shallow information in the text. For example, we observed that both sentence-level NMT and XLM-R produced a cluster responsible for extremely short sentences (the average sentence length is about four tokens for these clusters). On the other hand, document-level representations factor out these shallow stylistic features by averaging over sentence representations. Therefore, the models are inclined to cluster by topics. An alternative explanation is that domain-specific lexical statistics, which not all sentences might preserve, get more robust as we average sentence embeddings to get a document embedding.

Even though sentence-level clustering maintains a general idea about oracle domains, they split sentences into clusters quite freely. For example, JRC-Acquis consistently gets mixed with Europarl, which both belong to legal domains. We can see it from \textit{NMT SENT} for both language pairs.

For documents, the rediscovery trend is common and pronounced for both language pairs, and separation is generally consistent between train and test. However, for \textit{EN-ET XLM-R DOC} we can observe that EMEA and JRC-Acquis got split between two clusters in the training set. Considering that we perform ten random k-means restarts and choose the best iteration, this suggests that XLM-R may become inconsistent (as a source of sentence representations on the document level) in some cases. 

Figure \ref{fig:conf_tables-de_en} in Appendix \ref{app:analysis} shows similar heatmaps for DE-EN. DE-EN is consistent with what we observe for EN-ET apart from XLM-R DOC, where the DE-EN diagonal is cleaner.

\section{Practical Application} 
\label{sec:framework}
Our analysis in Section \ref{sec:analysis} revealed that NMT models represent domains in their embedding space separately, similarly to what pre-trained language models do \cite{aharoni-goldberg-2020-unsupervised}. We demonstrated that simple clustering on NMT representations allows recovering original data domains to a large degree. 

This section proposes to utilize this finding to improve an existing framework of automatic domain generation for NMT. In this framework, related work first clusters the training data using representations from an external encoder, and then the baseline NMT model is adapted (fine-tuned) on each cluster separately. We propose to re-use the NMT baseline itself as the encoder in this framework.

Representations extracted from translation Transformers are specific to the task of translation. We hypothesise that it might result in clusters most suitable for downstream translation tasks like fine-tuning to specific domains/clusters.

Moreover, an advantage of our scenario is that we cluster the same data (with our NMT model) that we use for NMT model training. It is a frequent multi-domain NMT setup, where multiple target domains are available in training.  In the pre-trained language model setup, the data will be more out-of-domain, despite the model's generality.

\subsection{Existing Framework (Background)}
\label{sec:framework:existing}
In this subsection we describe an existing framework which uses automatic domains (clusters) to perform NMT domain adaptation \cite{tars18}. Recent work \cite{currey-etal-2020-distilling} employs large pretrained language models as part of the framework. It consists of several steps.

In step 1.1, we begin with a single heterogeneous dataset ("Original Dataset") and train a baseline NMT model on it. At the same time (step 1.2), we pass this dataset through the external pre-trained XLM-R model to extract hidden sentence/document representations for the whole dataset. In step 2,  we use the extracted sentence/document representations to train a k-means clustering model. In step 3, we use this k-means model to separate the original dataset into sub-datasets corresponding to the clusters. Lastly, we use the cluster-specific datasets to fine-tune the baseline NMT model from step 1.1 on each dataset separately, resulting in a set of specialized models. We use the k-means model at runtime to determine which NMT model to use to translate a new sentence/document. If we only use sentence clusters, the approach is equivalent to the one proposed by \citet{currey-etal-2020-distilling}. Refer to Figure \ref{fig:method:xlmr} from Appendix \ref{app:analysis} for the illustration of the steps described above.

\subsection{Improved Framework (Ours)}
\label{sec:framework:ours}

In this subsection we describe our modification to the existing automatic domains pipeline presented in Section \ref{sec:framework:existing}.

We propose reusing an NMT baseline to produce sentence representations for the clustering step instead of using an external encoder. Specifically, in step 1, we train a baseline NMT model just like in the existing framework. However, we found we can omit using the XLM-R model (step 1.2). Instead, to extract sentence/document representations for step 2, we reuse the trained NMT baseline. The rest of the pipeline remains the same. Figure \ref{fig:method:nmt} (Appendix \ref{app:analysis}) illustrates the updated framework.

Moreover, to produce clusters in both frameworks, we additionally study text representations on the level of documents.

\section{Experiments}
\label{sec:experiments}

In this section we perform an extensive experimental study comparing performance of the existing automatic domains framework (Section \ref{sec:framework:existing}) with our proposed version (Section \ref{sec:framework:ours}). We experiment with both sentence-level and document-level representations as a basis for k-means algorithm on three language pairs and two data scenarios.

We first train baseline Transformer NMT models on concatenated data from all domains (same baseline as in Section \ref{sec:analysis}) and then cluster the training, development, and test data using either this same baseline or XLM-R. Next, we fine-tune\footnote{We use the terms "fine-tuning" and "NMT domain adaptation" interchangeably.} our baseline models to the different obtained data partitions (clusters) and compare the translation quality of resulting fine-tuned (adapted) models.

\subsection{Setup}
We explore two data scenarios. First, we perform experiments on a mixture of distinct corpora. For these experiments, we reuse the data and \textit{concat} baseline NMT model (Transformer-base) described in Section \ref{sec:analysis} (EN-ET and DE-EN). In this setting, we can compare the performance of models fine-tuned to automatically discovered domains to that of oracle models (fine-tuned using known domains/datasets). We also randomly partition the data (into equal parts) and fine-tune the baseline models to them to get our lower bound estimates.

Second, we explore a scenario with a single corpus, which is highly heterogenous, and thus may contain multiple domains which are unknown. In this setting, we use the ParaCrawl \citep{paracrawl} parallel corpus\footnote{https://paracrawl.eu/}, which consists of diverse documents crawled from the web. We use three language pairs: English$\rightarrow$Estonian (EN-ET), German$\rightarrow$English (DE-EN), and English$\rightarrow$Czech (EN-CS). We use  $\sim$3M sentence pairs for all languages for training, and $\sim$3,000 sentence-pairs for  development and testing. The exact experimental setup with data sizes, training and preprocessing details can be found in in Appendix \ref{app:experiments_setup}.

% We train a baseline NMT model for 60 epochs on data from all domains (concat) similarly to baseline in  and then fine-tune it to different known domains (corpora) or unknown domains (clusters). %initialize the parameters of models fine-tuned to different known domains (corpora) or unknown domains (clusters). 

For our \textit{concat} baselines we follow the setup from Section \ref{sec:analysis:setup} (described in more detail in Appendix \ref{app:experiments_setup}). In the mixture of corpora experiments, baseline fine-tuning is performed for 50 epochs, and in the single heterogenous corpus experiments for 25 epochs (fine-tuning hyperparameters can be found in Appendix \ref{app:experiments_setup}). 
%Fine-tuning was performed with initial learning rate $1.25\times10^{-4}$, reducing by a factor of 0.5 every time the development loss has not improved for 3 consecutive epochs. 
For comparison, we also continue training the baseline models for longer as suggested by \citet{gururangan-etal-2020-dont} (\textit{concat-cont}). We continue training for the same number of epochs fine-tuning is done for in the corresponding experiment. %\todo[]{also mention random FT lower bound}

For each of the models, we evaluate the checkpoint which shows the highest BLEU score on the particular model's development set, and translate the test sets with beam size set to 5. We use the BLEU score \citep{papineni2002bleu}, specifically, the sacreBLEU implementation \citep{post2018call} to assess the models' translation performance. To test for statistical significance, we use paired bootstrap resampling \citep{koehn-significance-tests}.

\subsection{Labelled Domain Mix Experiments}
\label{sec:multi_domain}
 
%\todo[]{rewrite this?}
In this section we consider a scenario which can be practically interesting in cases where the data consists of several distinct domains with the labels unavailable or corrupted as in \citet{currey-etal-2020-distilling}. Moreover, it serves as an oracle experiment showing how well automatic domains perform compared to the golden labels. This way we have a better idea what to expect when applying them to unlabeled data as in Section \ref{sec:heterogeneous}. 
 
Table \ref{tab:de-en-concat-no-delta} shows the results for DE$\rightarrow$EN. We see that, for all corpora except Europarl, at least one model of the two that are based on document-level clustering always manages to surpass the oracle performance obtained by fine-tuning to known domains, and on Europarl the document-level models perform comparably to oracle. In most cases, document-level models show significantly better translation quality than XLM-R sentence-level models, which have been used in previous work, while NMT sentence-level models closely match the performance of XLM-R sentence ones. When scores are averaged over all four domains, document clustering obtained from the NMT encoder is the overall winner.

\begin{table}[]
\centering
\footnotesize
\begin{tabular}{@{}llllll@{}}
\toprule
% & \multicolumn{5}{c}{BLEU}
% \\\cmidrule(lr){2-6}
 & \multicolumn{1}{l}{EP} & \multicolumn{1}{l}{OS} & \multicolumn{1}{l}{JRC} & \multicolumn{1}{l}{EMEA} & \multicolumn{1}{l}{avg} \\ \midrule
concat & 37.2 & 21.7 & 52.3 & 73.8 & 46.25 \\
concat-cont & 37.2 & 22.3 & 52.4 & 73.7 & 46.40 \\
oracle & \textbf{37.4} & 22.6 & 53.4 & 74.7 & 47.03 \\ \midrule
sentence \\
\hspace{3mm} XLM-R & 36.6 & 22.4 & 52.8 & 73.9 & 46.43 \\
\hspace{3mm} NMT & 36.6 & 22.3 & 52.8 & 74.0 & 46.43 \\
document \\
\hspace{3mm} XLM-R & 37.3** & \textbf{22.9}* & 53.0 & 75.0** & 47.05 \\
\hspace{3mm} NMT & 37.3** & 22.5 & \textbf{53.7}** & \textbf{75.4}** & \textbf{47.23} \\ \midrule
random & 36.8 & 22.4 & 51.7 & 73.3 & 46.05 \\ \bottomrule
\end{tabular}
\caption{\label{tab:de-en-concat-no-delta}BLEU scores of the DE-EN baseline models, models fine-tuned to known corpora (oracle), to the proposed automatic domains, and to a random partitioning of the data. EP, JRC, EMEA and OS stand for Europarl, JRC-Acquis, EMEA and OpenSubtitles test sets, respectively. Statistically significant improvements of our proposed methods over sentence-level XLM-R clustering are marked with * ($p\leq0.05$) or ** ($p\leq0.01$). Document-level clustering matches and slightly surpasses the performance of fine-tuning on oracle domains. %Note that document-based clusters highly correlate with oracle domains themselves.
}
\end{table}

%Both NMT and XLM-R models perform comparably, which suggests stability of document-level clustering.

Table \ref{tab:en-et-concat-no-delta} shows results for the EN$\rightarrow$ET language pair. While fine-tuning on oracle domains yields an average improvement of 0.8 BLEU points over the baseline, fine-tuning on unsupervised document clusters obtained from the NMT encoder allows us to match that performance. However, for the EMEA test set XLM-R sentence clusters turn out to be the most successful approach, showing significantly higher BLEU scores than all other automatic partitions and outperforming the oracle by 1.2 BLEU points, while document-level NMT clustering also manages to surpass the oracle performance, albeit slightly. For OpenSubtitles and JRC-Acquis, oracle shows the highest overall scores, with document-level NMT clustering a close second, %\todo[]{do significance tests for baseline?} 
outperforming XLM-R sentence clustering by a noticeable margin. For OpenSubtitles, however, none of the automatic domain approaches manage to improve the baseline performance (and neither does continued training of the baseline), and even the oracle partition does not manage to do so by a statistically significant degree. %\todo[]{do i need more significance tests :(}
For Europarl, all automatic domain approaches yield comparable BLEU scores, with none being significantly better or worse than XLM-R sentence clusters.
      
Document-level XLM-R automatic domains have a low average score due to underperforming on the EMEA test set. We see from Figure \ref{fig:conf_tables} that this is a case of train-test mismatch: the EMEA test set is mostly translated by the model fine-tuned on cluster 1, whose training set predominantly consists of Europarl data. Cluster 0, which sees the most EMEA examples during fine-tuning, is not used to translate the test set at all, as we see from Figure \ref{fig:conf_tables}.

\begin{table}[]
\centering
\footnotesize
\begin{tabular}{@{}llllll@{}}
\toprule
% & \multicolumn{5}{c}{BLEU}
% \\\cmidrule(lr){2-6}

 \multicolumn{1}{l}{} & \multicolumn{1}{l}{EP} & \multicolumn{1}{l}{OS} & \multicolumn{1}{l}{JRC} & \multicolumn{1}{l}{EMEA} & \multicolumn{1}{l}{avg} \\ \midrule
concat & 28.7 & 19.1 & 47.3 & 47.8 & 35.725 \\
concat-cont & 28.8 & 18.5 & 48.4 & 48.5 & 36.050 \\
oracle & 28.7 & \textbf{19.2} & \textbf{50.0} & 48.2 & 36.525 \\ \midrule
sentence \\
\hspace{3mm} XLM-R & 29.0 & 18.6 & 48.9 & \textbf{49.4} & 36.475 \\
\hspace{3mm} NMT & 29.1 & 18.7 & 49.0 & 48.1$^{\dagger\dagger}$ & 36.225 \\
document \\
\hspace{3mm} XLM-R & \textbf{29.2} & 18.6 & 47.9$^{\dagger\dagger}$ & 39.2$^{\dagger\dagger}$ & 33.725 \\
\hspace{3mm} NMT & 29.1 & 19.0* & 49.8** & 48.4$^{\dagger\dagger}$ & \textbf{36.575} \\ \midrule
random & 28.5 & 18.5 & 47.1 & 47.0 & 35.275 \\ \bottomrule
\end{tabular}
\caption{\label{tab:en-et-concat-no-delta}
BLEU scores of the EN-ET baseline models, models fine-tuned to known corpora (oracle), to the proposed automatic domains, and to a random partitioning of the data. EP, JRC, EMEA and OS stand for Europarl, JRC-Acquis, EMEA and OpenSubtitles test sets, respectively. Statistically significant improvements of our proposed methods over sentence-level XLM-R clustering are marked with * ($p\leq0.05$) or ** ($p\leq0.01$), daggers mark results which are significantly lower than for sentence-level clustering based on XLM-R ($\dagger$ and $\dagger\dagger$ denote $p\leq0.05$ and $p\leq0.01$, respectively). Document-level clustering as well as XLM-R based sentence-level clustering match the performance of the fine-tuning on oracle domains. %Note that clusters obtained from document based clustering highly correlate with oracle domains themselves.
}
\end{table}

\subsection{Heterogeneous Corpus Experiments}
\label{sec:heterogeneous}

In this subsection we present results for our method applied to the Paracrawl dataset, which constitutes a heterogeneous corpus of data crawled from the web with no training-time domain information known.

\paragraph{EN-ET}

\begin{table}[]
\centering
\small
\begin{tabular}{@{}lllll@{}}
\toprule
N of clusters & 3 & 4 & 5 & 8 \\  \midrule
concat & 46.1 & 46.1 & 46.1 & 46.1 \\
concat-cont & 46.6 & 46.6 & 46.6 & 46.6 \\ \midrule
sentence \\
\hspace{3mm} XLM-R & \textbf{47.0} & 46.8 & 47.1 & 47.0\\
\hspace{3mm} NMT & 46.9 & 47.1 & \textbf{47.6}** & 47.4* \\
document \\
\hspace{3mm} XLM-R & 46.8 & \textbf{47.2}* & 47.3 & 47.6**\\
\hspace{3mm} NMT & 46.8 & 47.0 & 47.2 & \textbf{48.2}**\\ \midrule
random & 46.1 & 45.9 & 45.5 & 45.3\\ \bottomrule
\end{tabular}
\caption{BLEU scores of models trained on EN-ET ParaCrawl and fine-tuned to different numbers of automatic clusters and to a random partitioning of the data. Statistically significant improvements of our proposed methods over sentence-level XLM-R clustering are marked with * ($p\leq0.05$) or ** ($p\leq0.01$). For different numbers of clusters different approaches score best, but the best result overall is obtained with document-level NMT and 8 clusters.}  % Our approach with \textit{document level} clustering, where \textit{NMT model} is used to extract representations, outperforms baseline and all other clustering approaches when the number of clusters is high enough.}
\label{tab:en-et-paracrawl}
\end{table}

We first experiment on the EN-ET language pair. While in the multi-corpus setup we chose the number of clusters to match the number of different corpora in our training set, in the ParaCrawl experiments we do not have a predefined number of domains. Therefore, we experiment with separating the dataset into 3, 4, 5, and 8 clusters.

The resulting BLEU scores for EN-ET are shown in Table \ref{tab:en-et-paracrawl}. Fine-tuning based on NMT and XLM-R clustering of the data outperforms a strong \emph{concat-cont} baseline by 0.2-1.6 BLEU points depending on the choice of embedding model and clustering level. The best result overall is achieved by our document-level NMT clustering, which also outperforms all other approaches with 8 clusters by at least 0.6 BLEU. Both document-level approaches %consistently 
improve their performance with a growing number of clusters. With 3 clusters, all clustering methods show comparable results, with none being significantly better or worse than sentence-level XLM-R. Document-level XLM-R and sentence-level NMT significantly outperform sentence-level XLM-R with 4 and 5 clusters, respectively.

\paragraph{EN-CS \& DE-EN}

As separating the data into 8 clusters yields the highest BLEU score among all fine-tuning scenarios for EN$\rightarrow$ET, we choose this number of clusters for experiments on other language pairs. Table \ref{tab:all-3m-paracrawl} shows the BLEU scores for EN$\rightarrow$ET, EN$\rightarrow$CS, and DE$\rightarrow$EN models fine-tuned to automatic domains.

For EN-CS, only the NMT sentence-level clustering manages to outperform the baseline, noticeably surpassing all other automatic domain extraction methods as well.

For DE-EN, none of the approaches outperform the baseline model by a considerable margin. Sentence-level clustering based on XLM-R performs comparably to the baseline. Document-level NMT clustering shows a slightly lower score, but the difference is not statistically significant. At the same time, document XLM-R and sentence NMT perform worse than sentence XLM-R.

%\todo[inline]{more description? + this needs some conclusion}

\begin{table}[]
\centering
\small
\begin{tabular}{@{}lllll@{}}
\toprule
 & EN-ET & EN-CS & DE-EN \\  \midrule
concat & 46.1 & 44.4 & 48.2 \\
concat-cont & 46.6 & 44.3 & 48.1 \\ \midrule
sentence \\
\hspace{3mm} XLM-R & 47.0 & 44.2 & \textbf{48.3} \\
\hspace{3mm} NMT & 47.4* & \textbf{44.9}** & 48.0$^{\dagger}$ \\
document \\
\hspace{3mm} XLM-R & 47.6** & 44.2 & 47.9$^{\dagger\dagger}$\\
\hspace{3mm} NMT & \textbf{48.2}** & 44.2 & 48.0\\ \midrule
random & 45.3 & 43.6 & 47.5\\ \bottomrule
\end{tabular}
\caption{BLEU scores of models trained on ParaCrawl and fine-tuned to automatic clusters and to a random partitioning of the data on three language pairs. For each language pair we use \textasciitilde3M training examples and split the data into 8 clusters. Statistically significant improvements of our proposed methods over sentence-level XLM-R clustering are marked with * ($p\leq0.05$) or ** ($p\leq0.01$), Daggers mark results which are significantly lower than for sentence-level clustering based on XLM-R ($\dagger$ and $\dagger\dagger$ denote $p\leq0.05$ and $p\leq0.01$, respectively).}
\label{tab:all-3m-paracrawl}
\end{table}

% In summary, we can successfully use the clustering-based method to improve NMT quality for EN-ET language pair. We experimented with several approaches that involve XLM-R and NMT encoders. We considered clustering based on sentence and document levels. We showed that our novel NMT document-based clustering outperforms XLM-R counterparts in terms of BLEU on EN-ET language pair. Moreover, it also essentially wins in terms of efficiency because it does not require a large external language model. We show that one can reuse the \emph{concat} model to extract suitable representations from the training set.

\subsection{Additional Exploration}
\label{sec:additional}
While automatic domains demonstrate reasonable performance for EN-ET and EN-CS language pairs, DE-EN does not seem to benefit from either XLM-R or NMT-based clustering. In this section we perform additional experiments with DE-EN data to see whether there are conditions under which automatic domains could be beneficial in this case.

\paragraph{Data Size and Number of Clusters}
First, we increase the training data size and vary the number of clusters. Specifically, we use 10M parallel sentence pairs for training instead of 3M, and partition the dataset into 4 and 12 clusters instead of 8.

The resulting BLEU scores for DE-EN are shown in Table \ref{tab:de-en-bigdata}. We do not observe any significant improvement over the \textit{concat-cont} baseline for any of the methods. With the data separated into 12 clusters, sentence-level NMT clustering significantly outperforms sentence-level XLM-R, but still does not beat continued training of the baseline.

\begin{table}[]
    \centering
    \small
    \begin{tabular}{@{}lll@{}}
         \toprule
         N of clusters & 4 & 12 \\  \midrule
         concat & 50.6 & 50.6  \\
         concat-cont & 50.9 & 50.9  \\ \midrule
         sentence \\
         \hspace{3mm} XLM-R & 50.9 & 50.6 \\
         \hspace{3mm} NMT & 51.0 & \textbf{50.9}* \\
         document \\
         \hspace{3mm} XLM-R & \textbf{51.1} & 50.8 \\
         \hspace{3mm} NMT & \textbf{51.1} & 50.8 \\ \midrule
         random & 50.2 & 49.8 \\ \bottomrule
    \end{tabular}
    \caption{\label{tab:de-en-bigdata} BLEU scores of models trained on 10M sentence pairs from DE-EN ParaCrawl and fine-tuned to 4 and 12 automatic clusters. The data size is increased compared to the previous experiments, the NMT model size remains the same. We see an improvement in baseline performance, but no improvement in the performance of fine-tuned models.  Statistically significant improvements of our proposed methods over sentence-level XLM-R clustering are marked with * ($p\leq0.05$).}
\end{table}

\paragraph{Model Size}

It is also possible that NMT needs different model capacity for handling different language pairs, so we experiment with decreasing the model size.
We use the same number of layers, but decrease the width of the model (4 attention heads, embeddings of size 160, dimension of the feed-forward layer 320) so that the total number of parameters decreases five-fold. We compute NMT clusters based on the new, smaller baseline model. Our motivation for this is to understand whether automatic domains are not useful for DE-EN ParaCrawl at all, or could aid a weaker baseline.

\begin{table}[]
    \centering
    \small
    \begin{tabular}{@{}lll@{}}
         \toprule
         Model size & Base & Small \\  \midrule
         concat & 48.2 & 44.2  \\
         concat-cont & 48.1 & 44.8  \\ \midrule
         sentence \\
         \hspace{3mm} XLM-R & \textbf{48.3} & 45.2 \\
         \hspace{3mm} NMT & 48.0$^{\dagger}$ & \textbf{45.4} \\
         document \\
         \hspace{3mm} XLM-R & 47.9$^{\dagger\dagger}$ & 45.1 \\
         \hspace{3mm} NMT & 48.0 & \textbf{45.4} \\ \midrule
         random & 47.5 & 44.0 \\ \bottomrule
    \end{tabular}
    \caption{\label{tab:de-en-smallmodel}BLEU scores of models trained on 3M sentence pairs from DE-EN ParaCrawl and fine-tuned to 8 automatic clusters. The Base NMT model has the same configuration as in previous experiments (\emph{Transformer-base}), while the Small model has 5 times fewer parameters. The smaller model benefits from fine-tuning to automatic domains, but does so starting from a weaker baseline performance. Daggers mark results which are significantly lower compared to sentence-level clustering based on XLM-R ($\dagger$ and $\dagger\dagger$ denote $p\leq0.05$ and $p\leq0.01$, respectively).}
\end{table}

The results are shown in Table \ref{tab:de-en-smallmodel}. The smaller baseline does benefit from adaptation to automatic domains (clusters). While NMT clusters are generated by a model which is 5 times as small, XLM-R and NMT show equivalent performance.

\section{Discussion}

Our analysis is implicit inductive evidence for the high degrees of domain-specific information in sentence and document NMT representations. However, it is still open to what kind of information is preserved (topical/stylistic/lexical).

For example, our approach could result in clusters by domain/dataset due to standard lexical statistics and not sentence semantics. However, on the practical side, we show that adapting NMT to these types of clusters is just as good or better as to other possible types of clusters since it benefits the baseline performance. Moreover, previous work that uses pre-trained language models to obtain the clusters is likely to suffer from the same issue.

Moreover, while XLM-R is a general-purpose encoder, NMT models are only that helpful for domains we train them on. However, the data constitutes all domains of interest by definition for a multi-domain NMT (the task we tackle). Thus, NMT models are a perfect fit that simplifies and outperforms an existing approach.

\section{Conclusion}
In this work, we made a two-fold contribution. The first is to the field of NMT interpretation and analysis. We have shown that a baseline Transformer NMT encoder preserves enough domain-specific information to distinguish between oracle domains in a mixed corpus without supervision. We showed an evolution of this property across the Transformer layer using PCA and k-means clustering on the level of sentences and documents. Comparison to XLM-R based clusters demonstrated that both sentence-level and document-level NMT clusters show higher cluster purity (similarity to original text domains).%, whereas document-level clusters near 100\%.

Next, we utilized our analysis insights to improve an existing practical cluster-based multi-domain NMT approach \citep{tars18, currey-etal-2020-distilling}. In a setting with preset domains (i.e., available corpus/domain labels), tuning to NMT clusters on average matches or surpasses XLM-R clusters. Additionally, NMT cluster-based tuning mostly matches the translation quality when tuning to original corpus labels, with some exceptions that we also analyze and explain.

Finally, in the case of a heterogeneous corpus (ParaCrawl), the performance of fine-tuned NMT models depends on the number of clusters, language pairs, and other parameters. We see significant improvement for EN-ET and EN-CS translation when comparing XLM-R and NMT-based clusters (on both sentence and document levels). For DE-EN, the domain tuning results depend on the NMT model's capacity for learning each language pair's translation.

\section*{Acknowledgements}

This work has been supported by the grant No. 825303 (Bergamot\footnote{\url{https://browser.mt/}}) of European Union’s Horizon 2020 research and innovation program.
The authors also thank the Unversity of Tartu's High-Performance Computing Center 
for providing GPU computing resources \cite{https://doi.org/10.23673/ph6n-0144}.

\bibliography{anthology,custom,deliverable}

\begin{thebibliography}{24}
\expandafter\ifx\csname natexlab\endcsname\relax\def\natexlab#1{#1}\fi

\bibitem[{Aharoni and Goldberg(2020)}]{aharoni-goldberg-2020-unsupervised}
Roee Aharoni and Yoav Goldberg. 2020.
\newblock \href {https://doi.org/10.18653/v1/2020.acl-main.692} {Unsupervised
  domain clusters in pretrained language models}.
\newblock In \emph{Proceedings of the 58th Annual Meeting of the Association
  for Computational Linguistics}, pages 7747--7763, Online. Association for
  Computational Linguistics.

\bibitem[{Bahdanau et~al.(2015)Bahdanau, Cho, and
  Bengio}]{Bahdanau2015NeuralMT}
Dzmitry Bahdanau, {Kyung Hyun} Cho, and Yoshua Bengio. 2015.
\newblock Neural machine translation by jointly learning to align and
  translate.
\newblock 3rd International Conference on Learning Representations, ICLR 2015 ;
  Conference date: 07-05-2015 Through 09-05-2015.

\bibitem[{Barrault et~al.(2020)Barrault, Biesialska, Bojar, Costa-juss{\`a},
  Federmann, Graham, Grundkiewicz, Haddow, Huck, Joanis, Kocmi, Koehn, Lo,
  Ljube{\v{s}}i{\'c}, Monz, Morishita, Nagata, Nakazawa, Pal, Post, and
  Zampieri}]{barrault-etal-2020-findings}
Lo{\"\i}c Barrault, Magdalena Biesialska, Ond{\v{r}}ej Bojar, Marta~R.
  Costa-juss{\`a}, Christian Federmann, Yvette Graham, Roman Grundkiewicz,
  Barry Haddow, Matthias Huck, Eric Joanis, Tom Kocmi, Philipp Koehn, Chi-kiu
  Lo, Nikola Ljube{\v{s}}i{\'c}, Christof Monz, Makoto Morishita, Masaaki
  Nagata, Toshiaki Nakazawa, Santanu Pal, Matt Post, and Marcos Zampieri. 2020.
\newblock \href {https://www.aclweb.org/anthology/2020.wmt-1.1} {Findings of
  the 2020 conference on machine translation ({WMT}20)}.
\newblock In \emph{Proceedings of the Fifth Conference on Machine Translation},
  pages 1--55, Online. Association for Computational Linguistics.

\bibitem[{Bojanowski et~al.(2017)Bojanowski, Grave, Joulin, and
  Mikolov}]{bojanowski-etal-2017-enriching}
Piotr Bojanowski, Edouard Grave, Armand Joulin, and Tomas Mikolov. 2017.
\newblock \href {https://doi.org/10.1162/tacl_a_00051} {Enriching word vectors
  with subword information}.
\newblock \emph{Transactions of the Association for Computational Linguistics},
  5:135--146.

\bibitem[{Conneau et~al.(2020)Conneau, Khandelwal, Goyal, Chaudhary, Wenzek,
  Guzm{\'a}n, Grave, Ott, Zettlemoyer, and
  Stoyanov}]{conneau-etal-2020-unsupervised}
Alexis Conneau, Kartikay Khandelwal, Naman Goyal, Vishrav Chaudhary, Guillaume
  Wenzek, Francisco Guzm{\'a}n, Edouard Grave, Myle Ott, Luke Zettlemoyer, and
  Veselin Stoyanov. 2020.
\newblock \href {https://doi.org/10.18653/v1/2020.acl-main.747} {Unsupervised
  cross-lingual representation learning at scale}.
\newblock In \emph{Proceedings of the 58th Annual Meeting of the Association
  for Computational Linguistics}, pages 8440--8451, Online. Association for
  Computational Linguistics.

\bibitem[{Currey et~al.(2020)Currey, Mathur, and
  Dinu}]{currey-etal-2020-distilling}
Anna Currey, Prashant Mathur, and Georgiana Dinu. 2020.
\newblock \href {https://doi.org/10.18653/v1/2020.emnlp-main.364} {Distilling
  multiple domains for neural machine translation}.
\newblock In \emph{Proceedings of the 2020 Conference on Empirical Methods in
  Natural Language Processing (EMNLP)}, pages 4500--4511, Online. Association
  for Computational Linguistics.

\bibitem[{Devlin et~al.(2019)Devlin, Chang, Lee, and
  Toutanova}]{devlin2019bert}
Jacob Devlin, Ming-Wei Chang, Kenton Lee, and Kristina Toutanova. 2019.
\newblock {{BERT}}: Pre-training of deep bidirectional transformers for
  language understanding.
\newblock In \emph{Proceedings of the 2019 Conference of the North American
  Chapter of the Association for Computational Linguistics: Human Language
  Technologies, Volume 1 (Long and Short Papers)}, pages 4171--4186.

\bibitem[{Espl{\`a} et~al.(2019)Espl{\`a}, Forcada, Ram{\'\i}rez-S{\'a}nchez,
  and Hoang}]{paracrawl}
Miquel Espl{\`a}, Mikel Forcada, Gema Ram{\'\i}rez-S{\'a}nchez, and Hieu Hoang.
  2019.
\newblock \href {https://www.aclweb.org/anthology/W19-6721} {{P}ara{C}rawl:
  Web-scale parallel corpora for the languages of the {EU}}.
\newblock In \emph{Proceedings of Machine Translation Summit XVII Volume 2:
  Translator, Project and User Tracks}, pages 118--119, Dublin, Ireland.
  European Association for Machine Translation.

\bibitem[{Gururangan et~al.(2020)Gururangan, Marasovi{\'c}, Swayamdipta, Lo,
  Beltagy, Downey, and Smith}]{gururangan-etal-2020-dont}
Suchin Gururangan, Ana Marasovi{\'c}, Swabha Swayamdipta, Kyle Lo, Iz~Beltagy,
  Doug Downey, and Noah~A. Smith. 2020.
\newblock \href {https://doi.org/10.18653/v1/2020.acl-main.740} {Don{'}t stop
  pretraining: Adapt language models to domains and tasks}.
\newblock In \emph{Proceedings of the 58th Annual Meeting of the Association
  for Computational Linguistics}, pages 8342--8360, Online. Association for
  Computational Linguistics.

\bibitem[{Hu et~al.(2020)Hu, Ruder, Siddhant, Neubig, Firat, and
  Johnson}]{hu2020extreme}
Junjie Hu, Sebastian Ruder, Aditya Siddhant, Graham Neubig, Orhan Firat, and
  Melvin Johnson. 2020.
\newblock \href {https://proceedings.mlr.press/v119/hu20b.html} {{XTREME}: A
  massively multilingual multi-task benchmark for evaluating cross-lingual
  generalisation}.
\newblock In \emph{Proceedings of the 37th International Conference on Machine
  Learning}, volume 119 of \emph{Proceedings of Machine Learning Research},
  pages 4411--4421. PMLR.

\bibitem[{Jiang et~al.(2020)Jiang, Liang, Wang, and
  Zhao}]{jiang-etal-2020-multi}
Haoming Jiang, Chen Liang, Chong Wang, and Tuo Zhao. 2020.
\newblock \href {https://doi.org/10.18653/v1/2020.acl-main.165} {Multi-domain
  neural machine translation with word-level adaptive layer-wise domain
  mixing}.
\newblock In \emph{Proceedings of the 58th Annual Meeting of the Association
  for Computational Linguistics}, pages 1823--1834, Online. Association for
  Computational Linguistics.

\bibitem[{Koehn(2004)}]{koehn-significance-tests}
Philipp Koehn. 2004.
\newblock Statistical significance tests for machine translation evaluation.
\newblock pages 388--395.

\bibitem[{Koehn(2005)}]{koehn-europarl}
Philipp Koehn. 2005.
\newblock {Europarl : A Parallel Corpus for Statistical Machine Translation}.
\newblock \emph{MT Summit}, 11.

\bibitem[{Kudo and Richardson(2018)}]{kudo2018sentencepiece}
Taku Kudo and John Richardson. 2018.
\newblock \href {https://doi.org/10.18653/v1/D18-2012} {{S}entence{P}iece: A
  simple and language independent subword tokenizer and detokenizer for neural
  text processing}.
\newblock In \emph{Proceedings of the 2018 Conference on Empirical Methods in
  Natural Language Processing: System Demonstrations}, pages 66--71, Brussels,
  Belgium. Association for Computational Linguistics.

\bibitem[{Lison and Tiedemann(2016)}]{opensubs}
Pierre Lison and Jörg Tiedemann. 2016.
\newblock Opensubtitles2016: Extracting large parallel corpora from movie and
  tv subtitles.
\newblock In \emph{Proceedings of the Tenth International Conference on
  Language Resources and Evaluation (LREC 2016)}, Paris, France.

\bibitem[{Ott et~al.(2019)Ott, Edunov, Baevski, Fan, Gross, Ng, Grangier, and
  Auli}]{fairseq}
Myle Ott, Sergey Edunov, Alexei Baevski, Angela Fan, Sam Gross, Nathan Ng,
  David Grangier, and Michael Auli. 2019.
\newblock fairseq: A fast, extensible toolkit for sequence modeling.
\newblock In \emph{Proceedings of NAACL-HLT 2019: Demonstrations}.

\bibitem[{Papineni et~al.(2002)Papineni, Roukos, Ward, and
  Zhu}]{papineni2002bleu}
Kishore Papineni, Salim Roukos, Todd Ward, and Wei-Jing Zhu. 2002.
\newblock Bleu: a method for automatic evaluation of machine translation.
\newblock In \emph{Proceedings of the 40th annual meeting on association for
  computational linguistics}, pages 311--318. Association for Computational
  Linguistics.

\bibitem[{Post(2018)}]{post2018call}
Matt Post. 2018.
\newblock A call for clarity in reporting bleu scores.
\newblock In \emph{Proceedings of the Third Conference on Machine Translation:
  Research Papers}, pages 186--191.

\bibitem[{Steinberger et~al.(2006)Steinberger, Pouliquen, Widiger, Ignat,
  Erjavec, Tufi{\c{s}}, and Varga}]{steinberger-jrc-acquis}
Ralf Steinberger, Bruno Pouliquen, Anna Widiger, Camelia Ignat, Tomaž Erjavec,
  Dan Tufi{\c{s}}, and Dániel Varga. 2006.
\newblock {The JRC-Acquis: A multilingual aligned parallel corpus with 20+
  languages}.
\newblock In \emph{Proceedings of the 5th International Conference on Language
  Resources and Evaluation, LREC 2006}.

\bibitem[{Tars and Fishel(2018)}]{tars18}
Sander Tars and Mark Fishel. 2018.
\newblock Multi-domain neural machine translation.
\newblock In \emph{Proceedings of EAMT}, pages 259--268, Alicante, Spain.

\bibitem[{Tiedemann(2012)}]{OPUS}
Jörg Tiedemann. 2012.
\newblock Parallel data, tools and interfaces in opus.
\newblock In \emph{Proceedings of the Eight International Conference on
  Language Resources and Evaluation (LREC'12)}, Istanbul, Turkey. European
  Language Resources Association (ELRA).

\bibitem[{{University of Tartu}(2018)}]{https://doi.org/10.23673/ph6n-0144}
{University of Tartu}. 2018.
\newblock \href {https://doi.org/10.23673/PH6N-0144} {Ut rocket cluster,
  https://doi.org/10.23673/ph6n-0144}.

\bibitem[{Vaswani et~al.(2017{\natexlab{a}})Vaswani, Shazeer, Parmar,
  Uszkoreit, Jones, Gomez, Kaiser, and Polosukhin}]{NIPS2017_3f5ee243}
Ashish Vaswani, Noam Shazeer, Niki Parmar, Jakob Uszkoreit, Llion Jones,
  Aidan~N Gomez, \L~ukasz Kaiser, and Illia Polosukhin. 2017{\natexlab{a}}.
\newblock \href
  {https://proceedings.neurips.cc/paper/2017/file/3f5ee243547dee91fbd053c1c4a845aa-Paper.pdf}
  {Attention is all you need}.
\newblock In \emph{Advances in Neural Information Processing Systems},
  volume~30. Curran Associates, Inc.

\bibitem[{Vaswani et~al.(2017{\natexlab{b}})Vaswani, Shazeer, Parmar,
  Uszkoreit, Jones, Gomez, Kaiser, and Polosukhin}]{vaswani2017attention}
Ashish Vaswani, Noam Shazeer, Niki Parmar, Jakob Uszkoreit, Llion Jones,
  Aidan~N Gomez, {\L}ukasz Kaiser, and Illia Polosukhin. 2017{\natexlab{b}}.
\newblock Attention is all you need.
\newblock In \emph{Advances in Neural Information Processing Systems}, pages
  5998--6008.

\end{thebibliography}
\bibliographystyle{acl_natbib}

\newpage
\clearpage
\appendix

\section{Analysis}
\label{app:analysis}

\subsection{Additional Figures}
\begin{figure*}[t!]
\includegraphics[width=\textwidth]{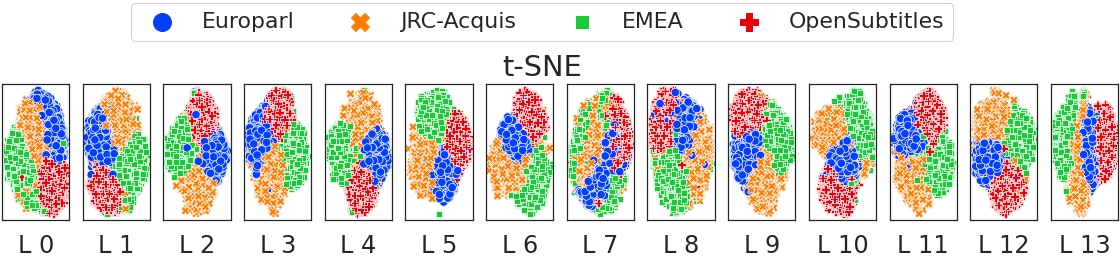}

\centering
\caption{%t-SNE and 
t-SNE plots of sentence representations extracted from all layers of the 60th checkpoint of the trained baseline NMT model. The dots, denoting sentences, are colored according to the domain the corresponding sentences come from. The model learns to distinguish between domains in its hidden space, despite not being explicitly provided with any information about domains. The figure shows that representations from the same domain cluster together.}

\label{fig:domains-tsne-only}
\end{figure*}

\begin{figure*}%
    \centering
    \subfloat[\centering DE-EN Train]{ {\includegraphics[width=\textwidth]{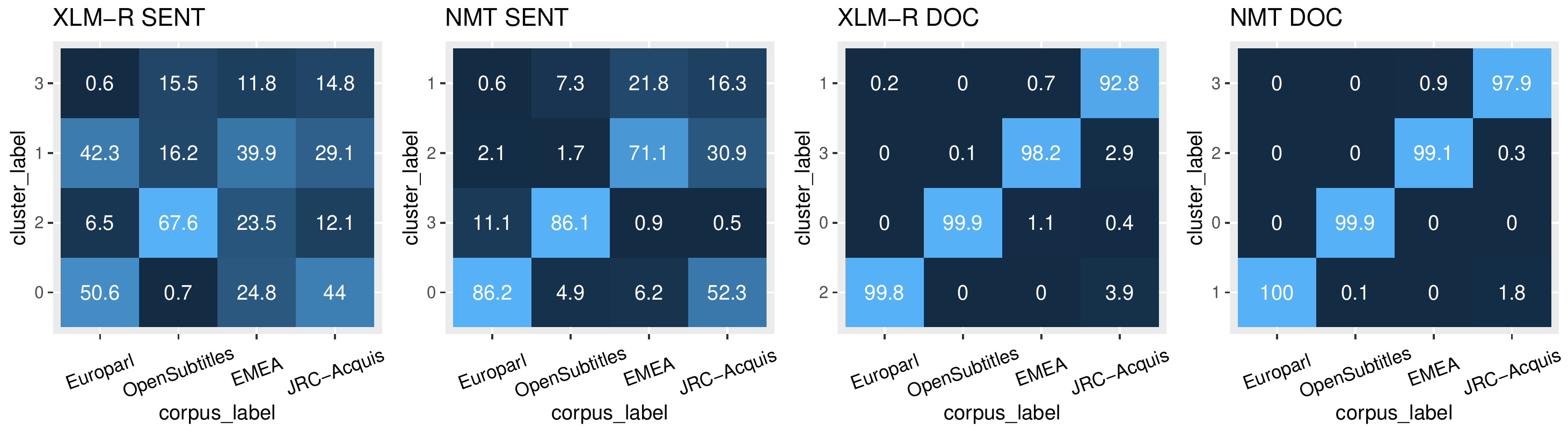} }}%
    \qquad
    \subfloat[\centering DE-EN Test]{ {\includegraphics[width=\textwidth]{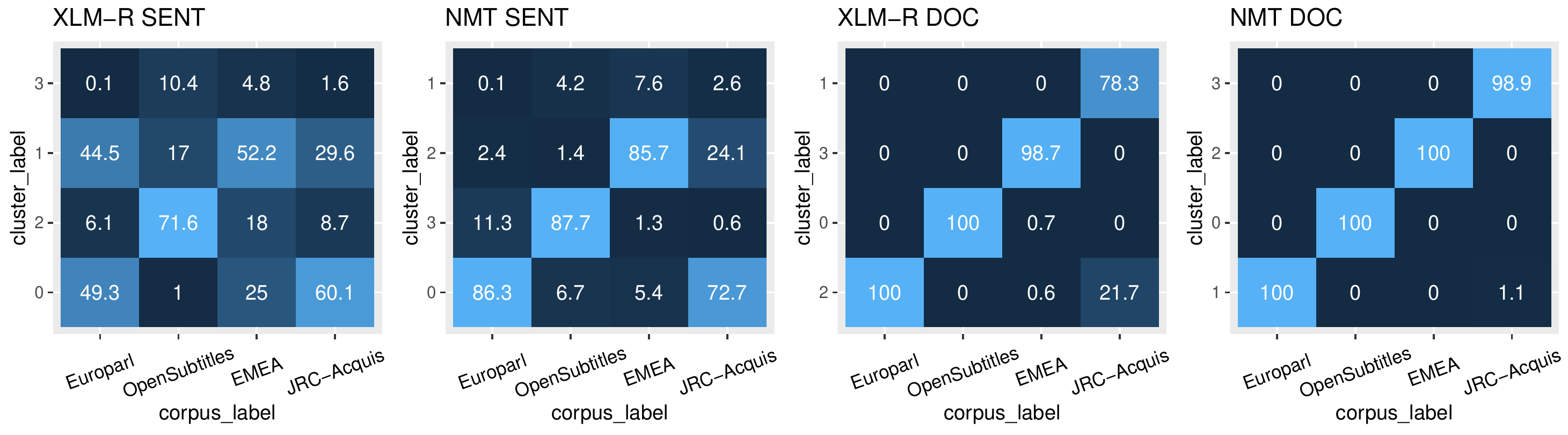} }}%
    \caption{Corpus-cluster confusion tables for about 2M sentences for DE-EN for the training (a) and test (b) sets. Document clusters rediscover original domains and NMT while sentence clusters tend to learn more customized notion of clusters. In general, NMT is more aligned to the oracle domains then BERT.}%
    \label{fig:conf_tables-de_en}%
\end{figure*}

\begin{figure}[H]
    \centering
    \includegraphics[width=\textwidth*4/9]{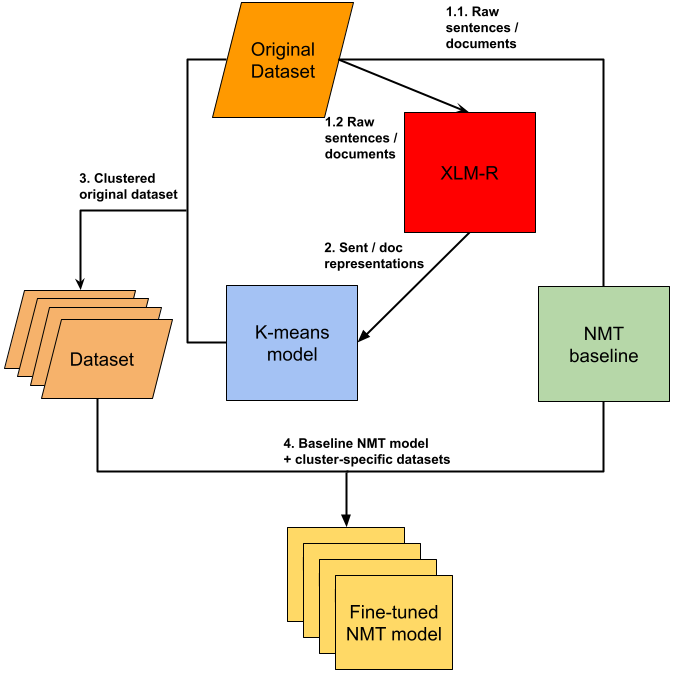}
    \caption{Existing automatic domains framework (previous approach).
    }
    \label{fig:method:xlmr}
\end{figure}

\begin{figure}[H]
    \centering
    \includegraphics[width=\textwidth*4/9]{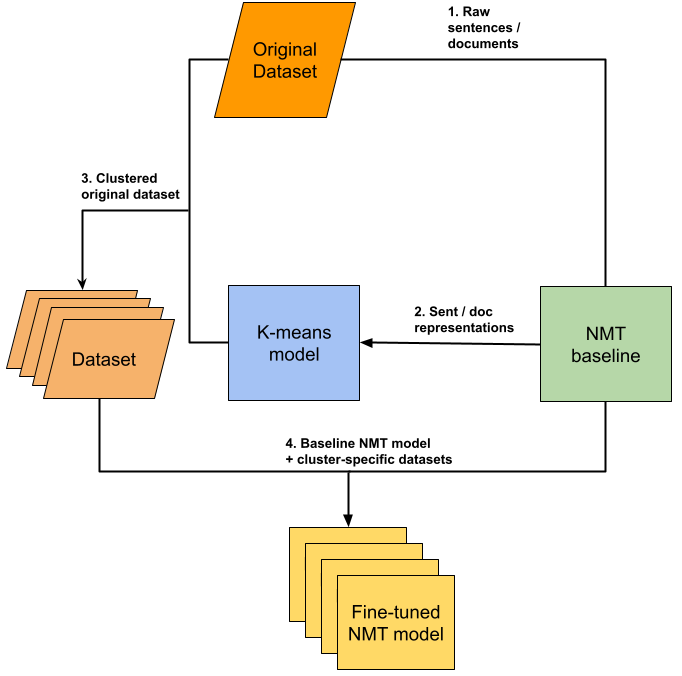}
    \caption{Updated automatic domains framework (ours).
    }
    \label{fig:method:nmt}
\end{figure}

Figures \ref{fig:domains-tsne-only} and \ref{fig:conf_tables-de_en} support our analysis in Section \ref{sec:analysis} while Figures \ref{fig:method:xlmr} and \ref{fig:method:nmt} illustrate frameworks from Section \ref{sec:framework}.

\subsection{Language Model}
\label{app:analysis:setup}

%\todo[inline]{we follow Currey et al. and...}

\paragraph{XLM-R Base} Our model of choice from the family of BERT-like models is the Base version of the XLM-R \citep{conneau-etal-2020-unsupervised}. It is a single multilingual model covering about 100 languages, which is very useful when dealing with machine translation systems, where for different language pairs we may not have a separate monolingual BERT for each source language. We choose XLM-R as opposed to the multilingual BERT \citep{devlin2019bert} since it is a more recent and better performing \citep{hu2020extreme} model. We choose the Base version because it is most compatible to our NMT baseline in terms of capacity.

\section{Experiments Setup}
\label{app:experiments_setup}

\subsection{Data}
\label{sec:experiments_setup:data}

For multi-domain fine-tuning (Section \ref{sec:multi_domain}) we experiment on German$\rightarrow$English (DE-EN) and English$\rightarrow$Estonian (ET-EN), and for the heterogenous corpus task (Section \ref{sec:heterogeneous}) we also evaluate on English$\rightarrow$Czech (EN-CS).

We use Europarl (proceedings of the European Parliament) \citep{koehn-europarl}, JRC-Acquis (legal documents of the European Union) \citep{steinberger-jrc-acquis}, EMEA (documents of the European Medicines Agency) \citep{OPUS} and OpenSubtitles (movie and TV subtitles) \citep{opensubs}\footnote{https://opus.nlpl.eu/} in the multidomain fine-tuning experiments. Data from the four corpora was approximately balanced. Around 500,000 training sentence pairs were taken from each of the corpora (except for EN-ET EMEA, where only 400,000 sentence pairs were available after cleaning), making the total size of the training set 1.9M sentence pairs for EN-ET and 2M for DE-EN. Development and test sets contain at least 3,000 sentences per corpus. The exact sizes of training, development and test sets can be found in Table \ref{tab:en-et-concat-datasize}. We test sentence-level and document-level clustering of the texts. For Europarl, JRC-Acquis, EMEA and OpenSubtitles, sentence pairs coming from the same XML file were considered to belong to the same document. The training, development and test sets in all experiments were constructed in such a way that a document is always included in one set in its entirety (hence the irregular sizes of the train, development and test sets). 

For the single heterogeneous corpus experiments we use v.7.1 of publicly available\footnote{https://paracrawl.eu/} Paracrawl dataset for all three language pairs. The training set sizes are 3M for all sentence pairs unless otherwise noted. Development and test sets contain at least 3,000 sentences per corpus in all experiments. The exact sizes of training, development and test sets in each of the experiments can be found in Tables \ref{tab:de-en-concat-datasize}, \ref{tab:en-et-concat-datasize}, and \ref{tab:paracrawl-datasize}. 
%We apply the same cleaning and preprocessing steps as before. 
In the ParaCrawl experiments, documents were matched by the source sentence URLs. The dataset is separated into webpage-based documents which we use to compile document-based clusters of representations produced by the baseline NMT and XLM-R models. The training, development and test sets in all experiments were constructed in such a way that a document is always included in one set in its entirety.

\begin{table*}[]
\centering
\begin{tabular}{@{}llllll@{}}
\toprule
 & Europarl & JRC-Acquis & OpenSubtitles & EMEA & total\\ \midrule
train & 500,697 (910) & 500,207 (8,874) & 501,510 (620) & 500,070 (941) & 2,002,484 (11,345)\\
dev & 3,566 (2) & 3,106 (78) & 4,306 (6) & 3,406 (10) & 14,384 (96) \\
test & 3,265 (12) & 3,008 (65) & 3,063 (3) & 5,908 (18) & 15,244 (98)\\ \bottomrule
\end{tabular}
\caption{\label{tab:de-en-concat-datasize} Number of sentence pairs (and documents) from each corpus (Europarl, JRC-Acquis, OpenSubtitles, EMEA) in the training, development and test sets of the DE-EN model trained on a mixture of known corpora}
\end{table*}

\begin{table*}[]
\centering
\begin{tabular}{@{}llllll@{}}
\toprule
 & Europarl & JRC-Acquis & OpenSubtitles & EMEA & total\\ \midrule
train & 500,166 (1,979) & 500,020 (8,877) & 500,876 (563) & 410,540 (732) & 1,911,602 (12,151)\\
dev & 3,716 (7) & 3,005 (91) & 3,044 (3) & 3,348 (10) & 13,113 (111)\\
test & 3,107 (16) & 3,190 (91) & 3,085 (4) & 3,315 (12) & 12,697 (123)\\ \bottomrule
\end{tabular}
\caption{\label{tab:en-et-concat-datasize} Number of sentence pairs (and documents) from each corpus (Europarl, JRC-Acquis, OpenSubtitles, EMEA) in the training, development and test sets of the EN-ET model trained on a mixture of known corpora}
\end{table*}

\begin{table*}[]
\centering
\begin{tabular}{@{}lllll@{}}
\toprule
 & EN-ET & EN-CS & DE-EN 3M & DE-EN 10M \\ \midrule
train & 3,163,124 (366,120) & 3,000,000 (777,448) & 3,000,013 (546,015) & 10,000,000 (1,819,571) \\
dev & 3,064 (400) & 3,019 (737) &3,018 (618) & 3,018 (618) \\
test & 3,130 (300) & 3,011 (770) & 3,007 (563) & 3,007 (563) \\ \bottomrule
\end{tabular}
\caption{\label{tab:paracrawl-datasize} Number of sentence pairs (and documents) in the training, development and test sets of the EN-ET, EN-CS, and DE-EN models trained on data from one heterogenous corpus (ParaCrawl)}
\end{table*}

Several basic cleaning steps were applied to the corpora. Sentence pairs were discarded if: 

\begin{itemize}
\item either the source or the target side was an empty string;
\item either the source or the target side contained more than 100 tokens;
\item one of the sentences in the pair contained at least 9 times as many tokens as the other;
\item more than half of the characters in either the source or the target sentence were non-alphabetic characters (noisy source or target)
\end{itemize}

In some corpora there are many sentence pairs that occur multiple times. Therefore, to avoid unfairly inflating the test scores, sentence pairs that also occur in the training set were removed from the development and test sets for BLEU score calculation in the multi-corpus experiments.

The data was split into subwords using SentencePiece \citep{kudo2018sentencepiece} with vocabulary size set to 32,000. No other pre-processing steps were applied.

\subsection{NMT Training}
\label{sec:experiments_setup:train_details}

We train Transformer machine translation models using the Fairseq toolkit \citep{fairseq}. The models have a standard configuration, mostly following the \textit{Transformer-base} settings \citep{NIPS2017_3f5ee243}: 6 encoder and 6 decoder layers, embedding dimension 512, feed-forward layer dimension 2048. The initial learning rate was set to $5\times10^{-4}$, with inverse square root learning rate scheduler with 4,000 warm-up updates. The loss function is label-smoothed cross entropy with label smoothing $\alpha$ equal to 0.1. We use Adam optimizer, with $\beta_1=0.9$ and $\beta_2=0.98$. Dropout probability is set to 0.3. The source and target vocabularies are shared. Model checkpoints are saved at the end of each epoch.

When fine-tuning, we pre-train the model without any explicit domain specific information, and then initialize each model with the parameters of the baseline's checkpoint from the 60th epoch. In the mixture of corpora experiments, fine-tuning is performed for 50 epochs, and in the single heterogenous corpus experiments for 25 epochs (our experiments show that for the overwhelming majority of models the checkpoint which has the best BLEU score on the development set occurs before 25 epochs of fine-tuning). Fine-tuning was performed with initial learning rate $1.25\times10^{-4}$, reducing by a factor of 0.5 every time the development loss has not improved for 3 consecutive epochs. For comparison, we also continue training the baseline model for the same number of epochs fine-tuning is done for. For each of the models, the translation is done with the checkpoint which has the highest BLEU score on the particular model's development set.

We use the BLEU score \citep{papineni2002bleu}, specifically, the sacreBLEU implementation \citep{post2018call} to assess the models' translation performance. To test for statistical significance, we use paired bootstrap resampling \citep{koehn-significance-tests}.

The models were pre-trained and fine-tuned either on one NVIDIA V100 GPU with 32GB of RAM with maximum batch size 15,000 tokens per node or on two NVIDIA V100 GPU's with 16GB of RAM with maximum batch size 7,500 tokens per node. The only exception is the DE-EN ParaCrawl model with 10M training sentence pairs, which has the largest volume of training data, and was pre-trained on 4 NVIDIA V100 GPU's with 32GB of RAM with maximum batch size 15,000 tokens per node.

\end{document}